\documentclass[10pt,twocolumn,letterpaper]{article}

\usepackage{cvpr}
\usepackage{times}
\usepackage{epsfig}
\usepackage{graphicx}
\usepackage{amsmath}
\usepackage{amssymb}
\usepackage{multirow}
\usepackage{epstopdf}
\usepackage{amssymb}
\usepackage{color}
\usepackage{float}
\usepackage{amsthm}
\usepackage[misc]{ifsym}

\usepackage[pagebackref=true,breaklinks=true,letterpaper=true,colorlinks,bookmarks=false]{hyperref}
\cvprfinalcopy 

\def\cvprPaperID{1724} 

\begin{document}

\title{Convolutional Neural Networks with Alternately Updated Clique}

\author{Yibo Yang$^{1,2}$,\quad Zhisheng Zhong$^2$,\quad Tiancheng Shen$^{1,2}$,\quad Zhouchen Lin$^{2,3,}$\thanks{Corresponding author}\\
\small $^1$Academy for Advanced Interdisciplinary Studies, Peking University\\
\small $^2$Key Laboratory of Machine Perception (MOE), School of EECS, Peking University\\
\small $^3$Cooperative Medianet Innovation Center, Shanghai Jiao Tong University\\
{\tt\small \{ibo,zszhong,tianchengShen,zlin\}@pku.edu.cn}
}

\maketitle
\thispagestyle{empty}

\begin{abstract}
\vspace{-2mm}
  Improving information flow in deep networks helps to ease the training difficulties and utilize parameters more efficiently. Here we propose a new convolutional neural network architecture with alternately updated clique (CliqueNet). In contrast to prior networks, there are both forward and backward connections between any two layers in the same block. The layers are constructed as a loop and are updated alternately. The CliqueNet has some unique properties. For each layer, it is both the input and output of any other layer in the same block, so that the information flow among layers is maximized. During propagation, the newly updated layers are concatenated to re-update previously updated layer, and parameters are reused for multiple times. This recurrent feedback structure is able to bring higher level visual information back to refine low-level filters and achieve spatial attention. We analyze the features generated at different stages and observe that using refined features leads to a better result. We adopt a multi-scale feature strategy that effectively avoids the progressive growth of parameters. Experiments on image recognition datasets including CIFAR-10, CIFAR-100, SVHN and ImageNet show that our proposed models achieve the state-of-the-art performance with fewer parameters
  \footnote{Code address: \url{http://github.com/iboing/CliqueNet}}.
\end{abstract}

\vspace{-4.5mm}

\section{Introduction}

In recent years, the structure and topology of deep neural networks have attracted significant research interests, since the convolutional neural network (CNN) based models have achieved huge success in a wide range of tasks of computer vision. A notable trend of those CNN architectures is that the layers are going deeper, from AlexNet \cite{krizhevsky2012imagenet} with 5 convolutional layers, the VGG network and GoogleLeNet with 19 and 22 layers, respectively \cite{simonyan2014very,szegedy2015going}, to recent ResNets \cite{he2016deep} whose deepest model has more than one thousand layers. However, inappropriately designed deep networks would make it hard for latter layer to access the gradient information from previous layers, which may cause gradient vanishing and parameter redundancy problems \cite{huang2016densely,huang2016deep}.


\begin{figure}[t]
   \includegraphics[width=1\linewidth]{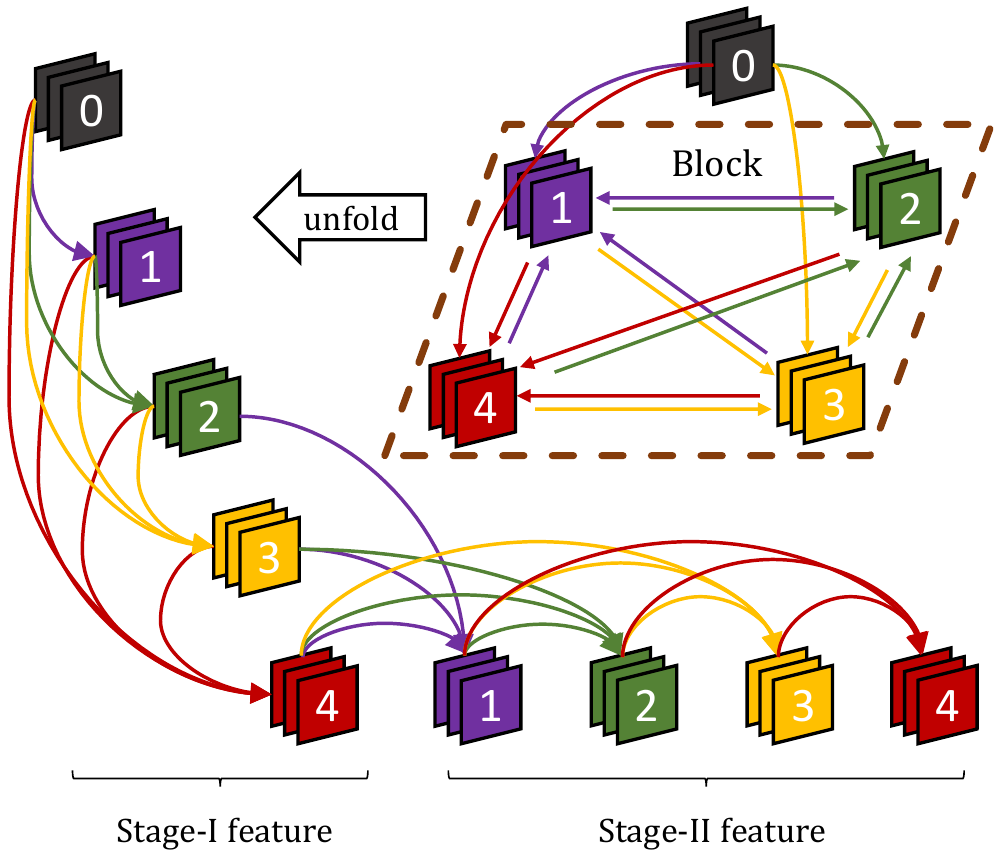}
   \caption{An illustration of a block with 4 layers. Any layer is both the input and output of another one. Node 0 denotes the input layer of this block.}
\label{fig:allblock}
\vspace{-5mm}
\end{figure}

Successfully adopted in ResNet \cite{he2016deep} and Highway Network \cite{srivastava2015training}, skip connection is an efficient way to make top layers accessible to the information from bottom layers, and ease the network training at the same time, due to its relief of the gradient vanishing problem. The residual block structure in ResNet \cite{he2016deep} also inspires a series of ResNet variations, including ResNext \cite{xie2016aggregated}, WRN \cite{zagoruyko2016wide}, PolyNet \cite{Zhang_2017_CVPR}, \emph{etc}. To further activate the gradient and information flow in networks, DenseNet \cite{huang2016densely} is a newly proposed structure, where any layer in a block is the output of all preceding layers, and the input of all subsequent layers. Recent studies show that the skip connection mechanism can be extrapolated as a recurrent neural network (RNN) or LSTM \cite{hochreiter1997long}, when weights are shared among different layers \cite{liao2016bridging,chen2017dual,jastrzebski2017residual}. In this way, the deep residual network is treated as a long sequence and hidden units are linked by skip connections. While this recurrent structure benefits feature re-usage and iterative learning, the residual information is restricted among neighboring layers and cannot be considered across multiple layers, because the recurrence only happens once at each single layer.

Attention mechanism is another focus of recent studies on network structure \cite{wang2014attentional,Wang_2017_CVPR,cao2015look,mnih2014recurrent} and applications \cite{chen2016sca,nam2016dual,kuen2016recurrent,fu2017look}. When people watch a picture or a scene, the information on our target is better captured if we re-look at or re-think the target with additional attention. In cognition theory, the activity of a neuron in visual cortex is influenced by other cortical area's responses transferred through feedback connections \cite{hupe1998cortical,hopfinger2000neural}. This motivates the introduce of feedback to deep networks \cite{stollenga2014deep,Zamir_2017_CVPR}. The feedback connections that bring back higher-level semantic information in a top-down manner are able to re-weight the focus, and suppress the non-relevant neuron activations of background and noises.

Inspired by the recurrent structure and attention mechanism, in this study, we propose a new convolutional neural network architecture with alternately updated clique (CliqueNet). In contrast to prior network structures, there are both forward and feedback connections between any two layers in the same block. As illustrated in Figure~\ref{fig:allblock}, the layers in Clique Block are constructed as a clique and are updated alternately. Concretely, the several previous layers are concatenated to update the next layer, after which, the newly updated layer is concatenated to re-update the previous layer, so that information flow and feedback mechanism can be maximized. Each layer in a block is both the input and output of another one, which means they are more densely connected than DenseNets \cite{huang2016densely}. We adopt a multi-scale feature strategy to compose the final representation with the block features in different map sizes.



CliqueNet architecture has some unique properties. An intuition would tell that our proposal is parameter-demanding, because given a block with $n$ layers, DenseNet \cite{huang2016densely} needs $C_n^2$ groups of parameters, while ours needs $A_n^2$ ($C$ and $A$ represents combination operator and permutation operator, respectively). However, the filters in DenseNet increase linearly as the depth rises \cite{chen2017dual}, which may leads to the rapid growth of parameters. In our architecture, only the Stage-II feature in each block is fed into the next block. It turns out that this is a more parameter-efficient way. In addition, traditional neural networks add a new layer with its corresponding parameters. As for CliqueNet, the weights among layers in a block keep recycling during propagation. The layers can be updated alternately for multiple times so that a deeper representation space is attained with the fixed number of parameters.

CliqueNet also shows a strong ability for representation learning due to the combination of recurrent structure and feedback mechanism. In each Clique Block, both forward and feedback are densely connected. The information flow is maximized and feature maps are repeatedly refined by attention. We show that our network architecture can suppress the activations of background and noises, and achieve competitive results without resorting to data augmentation.



The contributions in this study are listed as follows:

%
%
\vspace{-2.5mm}

\begin{itemize}
  \item We propose a new convolutional neural network architecture called CliqueNet, which incorporates both forward and backward connections between any two layers in the same block. The layers constructed as a loop are updated alternately. The CliqueNet that combines both recurrent structure and attention mechanism, is able to maximize information flow and achieve feature refinement. We show that the refined features are more discriminative and lead to a better performance.



%
      \vspace{-2.5mm}
  \item We adopt a multi-scale feature strategy that effectively circumvents the progressive increment of parameters, despite the extra feedback connections.
      \vspace{-2.5mm}
  \item We conduct experiments on four benchmark datasets including CIFAR-10, CIFAR-100, SVHN and ImageNet to demonstrate the superiority of our models.
\end{itemize}




\section{Related Work}

\begin{figure*}
   \center
   \includegraphics[width=0.96\linewidth]{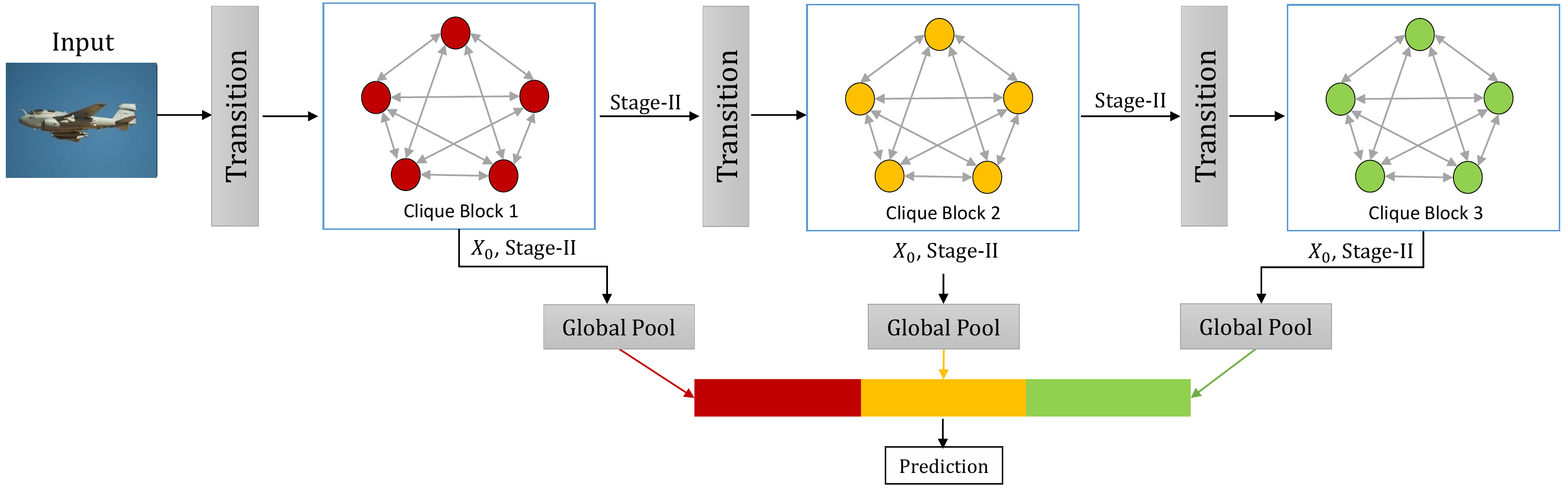}
   \caption{A CliqueNet with three blocks. The input layer together with the Stage-II feature in each block are concatenated to be the block feature, and form part of the final representation after global pooling. The Stage-II feature passes through transition layers, which include a convolution and an average pooling to change map sizes, and then becomes the input of the next block.}
\label{structure}
\vspace{-1.5mm}
\end{figure*}

A number of deep networks with large model capacity have been proposed. For widening the network, the Inception modules in GoogLeNet \cite{szegedy2015going} fuse the features in different map size to construct a multi-scale representation. Multi-column \cite{ciregan2012multi} nets and Deeply-Fused Nets \cite{wang2016deeply} also use fusion strategy and have a wide network structure. Wide residual networks \cite{zagoruyko2016wide} increase the width and decrease the depth to improve the performance, while FractalNet \cite{larsson2016fractalnet} deepen and widen at the same time. However, simply widening the network is easy to consume more runtime and memory \cite{Zhang_2017_CVPR}. For deepening the networks, skip connections or shortcut paths are widely adopted strategies to ease the network training \cite{he2016deep,srivastava2015training}. In \cite{huang2016deep}, it is shown that some of the layers in ResNets are dispensable and cause parameters redundancy. So they randomly drop a subset of layers to ease the training and achieve a better performance. To further increase information flow, DenseNets \cite{huang2016densely} replace the identity mapping in residual block by concatenating operation, so that new feature learning can be reinforced while keeping old feature re-usage. In line with this view, dual path networks (DPN) \cite{chen2017dual} are proposed to combine both advantages of residual path and densely connected path.

Both residual path and densely connected path correspond to a recurrent propagation, and their success has been attributed to the recurrent structure and iterative refinement \cite{liao2016bridging,greff2016highway,jastrzebski2017residual}. Studies incorporating recurrent connections into CNNs also show superiority in object recognition \cite{liang2015recurrent}, scene parsing \cite{Pinheiro2014Recurrent} and some other tasks. CliqueNet differs from these structures in that the iterative mechanism exists in each step of the propagation, instead of just between neighboring layers or from the top layer to the bottom layer; all layers in a block participate in the recurrent loop so that the filters are communicated sufficiently and the blocks play both roles of information carrier and refiner.

Recent studies have embraced the attention mechanism as an effective technique to strengthen some neurons that feature the target, and improve the performance as a result. It is proved fruitful in many applications, including image recognition \cite{Wang_2017_CVPR,fu2017look}, image captioning \cite{chen2016sca}, image-text matching \cite{nam2016dual}, and saliency detection \cite{kuen2016recurrent}. In general, visual attention can be achieved by formulating an optimization problem \cite{cao2015look}, weighting the activations spatially or channel-wisely \cite{chen2016sca,hu2017}, and introducing feedback connections \cite{wang2014attentional,stollenga2014deep,Zamir_2017_CVPR}. In \cite{Zamir_2017_CVPR}, the model makes consecutive decisions for a more accurate prediction via feedback connections. The input of the next decision is based on the output of the last decision. Experiments show that the top-down propagation is capable of refining lower-level features, and improving classification performance \cite{stollenga2014deep}, especially on datasets with noise and occlusion \cite{wang2014attentional,mnih2014recurrent}. But how to make a proper attention mechanism and boost the supervision between layers remains further exploration.

There are also some studies that design attention mechanism tied with recurrent neural networks \cite{mnih2014recurrent,kuen2016recurrent,fu2017look}. A recent report \cite{caswellloopy} tries to propose a loopy net, but it just repeats the skip connections and does not make layers communicated. The loopy inference adopted in \cite{pmlr-v37-chenb15, Zheng_2015_ICCV} shares a similar motivation with our work. However, they do not incorporate feedback connections, which are important for feature refinement. CliqueNet enables true cycling because of the alternate propagation. Although alternate updating has been an important method in the optimization theory \cite{gabay1976dual}, it has not been introduced into deep learning areas. At the best of out knowledge, we are the first to use updated layers to re-update previous layers alternately, and these layers construct a loop to cycle for multiple times.


\section{CliqueNet Architecture}

The CliqueNet architecture has two main ingredients, the block with alternately updated clique (Clique Block) to enable feature refinement, and the multi-scale feature strategy that facilitates parameter efficiency.

\subsection{Clique Block}

\begin{table*}[!ht]
\begin{center}
\begin{tabular}{c|c|c|c}  
\hline
Bottom Layers & Weights & Top Layer & Feature\\
\hline\hline
$X_0$ & $W_{01}$ & $X_1^{(1)}$ & \multirow{5}*{Stage-I}\\
\{$X_0,X_1^{(1)}$\} & \{$W_{02},W_{12}$\} & $X_2^{(1)}$ & ~\\
\{$X_0,X_1^{(1)},X_2^{(1)}$\} & \{$W_{03},W_{13},W_{23}$\} & $X_3^{(1)}$ & ~\\
\{$X_0,X_1^{(1)},X_2^{(1)},X_3^{(1)}$\} & \{$W_{04},W_{14},W_{24},W_{34}$\} & $X_4^{(1)}$ & ~\\
\{$X_0,X_1^{(1)},X_2^{(1)},X_3^{(1)},X_4^{(1)}$\} & \{$W_{05},W_{15},W_{25},W_{35},W_{45}$\} & $X_5^{(1)}$ & ~\\
\hline\hline
\{$X_2^{(1)},X_3^{(1)},X_4^{(1)},X_5^{(1)}$\} & \{$W_{21},W_{31},W_{41},W_{51}$\} & $X_1^{(2)}$ & \multirow{5}*{Stage-II}\\
\{$X_3^{(1)},X_4^{(1)},X_5^{(1)},X_1^{(2)}$\} & \{$W_{32},W_{42},W_{52},W_{12}$\} & $X_2^{(2)}$ & ~\\
\{$X_4^{(1)},X_5^{(1)},X_1^{(2)},X_2^{(2)}$\} & \{$W_{43},W_{53},W_{13},W_{23}$\} & $X_3^{(2)}$ & ~\\
\{$X_5^{(1)},X_1^{(2)},X_2^{(2)},X_3^{(2)}$\} & \{$W_{54},W_{14},W_{24},W_{34}$\} & $X_4^{(2)}$ & ~\\
\{$X_1^{(2)},X_2^{(2)},X_3^{(2)},X_4^{(2)}$\} & \{$W_{15},W_{25},W_{35},W_{45}$\} & $X_5^{(2)}$ & ~\\
\hline\hline
\multicolumn{4}{c}{$\cdots$}\\
\hline
\end{tabular}
\end{center}
\caption{A diagram of CliqueNet's propagation in a block with 5 layers. $W_{ij}$ is the weights of parameter from $X_i$ to $X_j$ and keeps re-used. ``\{\}'' denotes the concatenation operator. The Stage-II feature is to be transited as the input layer ($X_0$) of the next block.}
\label{propagation}
\end{table*}
In order to maximize the information flow among layers, we design the Clique Block. Any two layers in the same block are connected bidirectionally except for the input node. Compared with Dense Block \cite{huang2016densely} where each layer is the output of all previous layers, and the input of all subsequent layers, Clique Block makes each layer both the input and output of any other layers. The propagation of a Clique Block with 5 layers is illustrated in Table~\ref{propagation}. At the first stage, the input layer ($X_0$) initializes all layers in this block by single directional connections. Each updated layer is concatenated to update the next layer. From the second stage, the layers begin updating alternately. All layers except the top layer to be updated are concatenated as the bottom layer, and their corresponding parameters are also concatenated. Accordingly, the $i$th ($i\ge1$) layer in the $k$th ($k\ge2$) loop can be formulated as:
\begin{equation}
X_i^{(k)}=g\left(\sum\limits_{l<i}W_{li}*X_l^{(k)}+\sum\limits_{m>i}W_{mi}*X_m^{(k-1)}\right)
\end{equation}
where $*$ denotes the convolution operation with parameters $W$, and $g$ is the non-linear activation function. $W_{ij}$ keeps re-used in different stages. Each layer will always receive the feedback information from the layers that are updated more lately. It achieves a spatial attention mechanism due to the top-down refinement brought by each propagation. This recurrent feedback structure ensures that the communication is maximized among all layers in the block.

\subsection{Feature at Different Stages}


We analyze the features produced at different stages, and adopt a multi-scale feature strategy to avoid the rapid increment of parameters.

\begin{figure}[t]
   \includegraphics[width=1\linewidth]{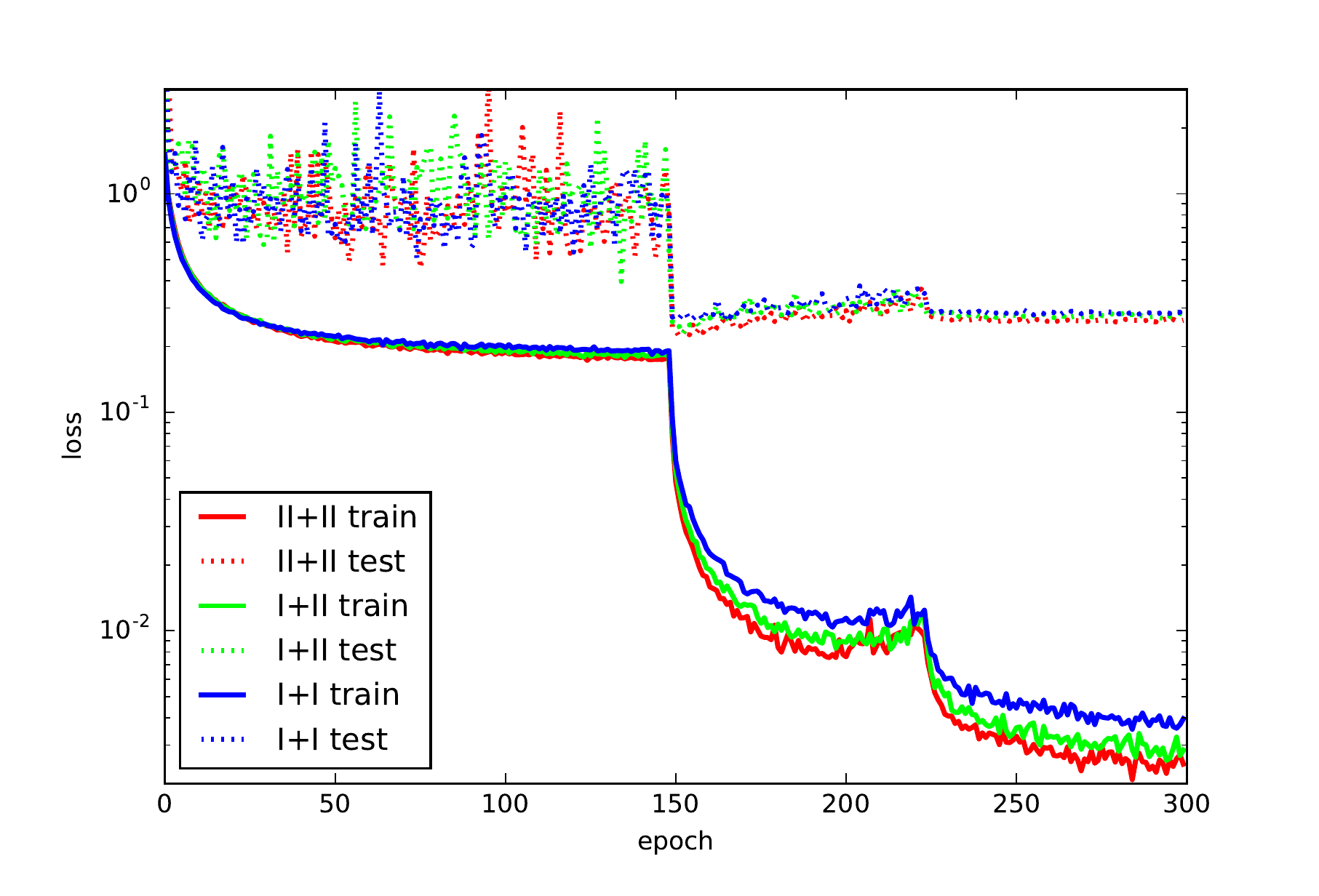}
   \caption{Training and testing curves of different versions of CliqueNets. Learning rate is divided by 10 at epoch 150 and 225.}
\label{fig:cifar-10}
\vspace{-1.8mm}
\end{figure}

The first stage is used to initialize all layers in the block, and the layers are refined repeatedly
since the second stage. Given that the Stage-II feature is refined with attention and assimilates more high level visual information, we make the Stage-II feature together with the input layer in each block concatenated as the block feature, and then accessed to the loss function after global pooling. Only the Stage-II feature is fed into the next block as their input layer $X_0$; see Figure~\ref{structure}. In this way, the final representation is characterized by multi-scale feature maps, and the dimensionality in each block will not increase progressively. Because higher stage propagation comes with more computational cost and amplifies the model complexity, we only consider the first two stages.

\begin{table}[ht]
\begin{center}
\begin{tabular}{c|c|c|c}  
\hline
name &block feature& transit & error(\%) \\
\hline\hline
CliqueNet (I+I) & $X_0$, Stage-I & Stage-I & 6.64 \\
CliqueNet (I+II) & $X_0$, Stage-I & Stage-II & 6.1 \\
CliqueNet (II+II) & $X_0$, Stage-II & Stage-II & 5.76 \\
\hline
\end{tabular}
\end{center}
\caption{Results of different versions of CliqueNets on CIFAR-10.}
\label{allnet-L}
\end{table}

For the purpose of analyzing the features generated in different stages, we conduct experiments on CIFAR-10 dataset (with no data augmentation) using different versions of CliqueNets. As Table~\ref{allnet-L} shows, the CliqueNet (I+I) only considers the Stage-I feature. The CliqueNet (I+II) uses the Stage-I feature and input layer as block feature to access loss function, but transits the Stage-II feature into the next block. The CliqueNet (II+II) adopts our aforementioned strategy. They all have 3 blocks with 5 layers in each block. Each layer contains 36 filters. The experimental settings are following \cite{huang2016densely}. The main results are shown in Figure~\ref{fig:cifar-10}. It is found that the introduce of Stage-II feature indeed leads to a better result by a significant margin. We adopt the CliqueNet (II+II) structure for the following experiments.

\subsection{Extra Techniques}

In addition to the structures mentioned above, we consider some techniques to help strengthen the model and improve the state of the art. In the experimental section, we conduct experiments with and without these additional techniques to show the effectiveness of our model.

\begin{figure}[t]
\begin{center}
   \includegraphics[width=0.9\linewidth]{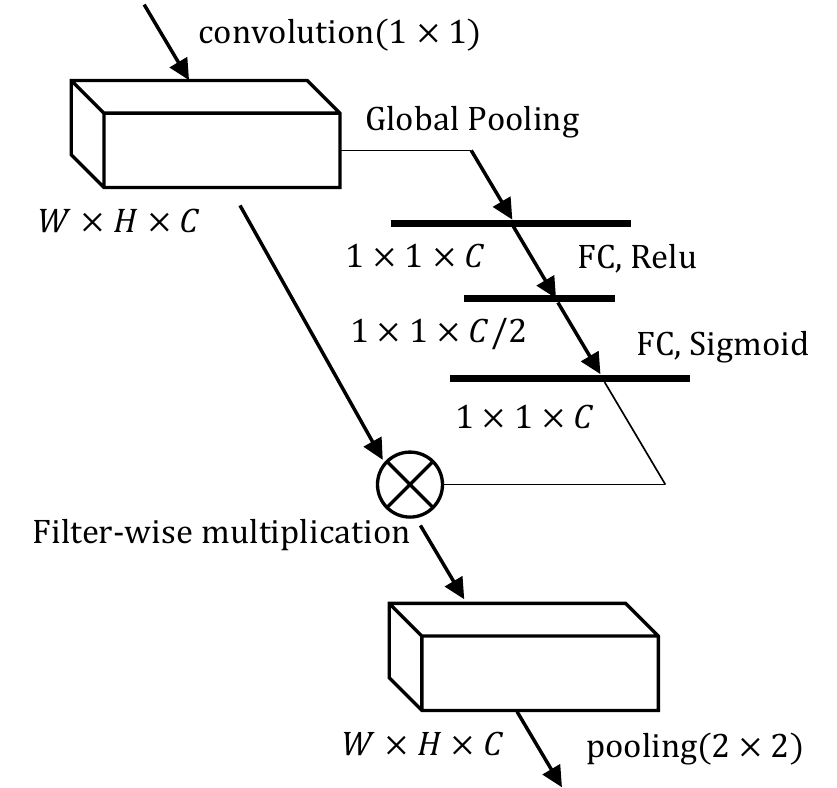}
\end{center}
   \caption{A schema for attentional transition. The transition layer consists of convolution and pooling. The filter-wise multiplication happens after convolution and before down pooling. $W$, $H$ and $C$ are width, height and channels of feature maps.}
\label{att-transition}
\end{figure}

\noindent\textbf{Attentional transition.} The CliqueNet includes feedback connections to refine lower level activations using higher level visual information. The attention mechanism weight the feature maps spatially to weaken the noises and background. The channel-wise attention, adopted in \cite{chen2016sca,Wang_2017_CVPR,hu2017}, also benefits recognition problem because it recalibrates different filters to prevent overfitting and inspire new features learning. In CliqueNet, we incorporate channel-wise attention mechanism in transition layers, following the method proposed in \cite{hu2017}. As depicted in Figure~\ref{att-transition}, the filters are globally averaged after the convolution in transition. They are followed by two fully connected (\emph{FC}) layers. The first \emph{FC} layer has half of the filters and is activated by \emph{Relu} function. The second \emph{FC} layer has the same number of filters and is activated by \emph{Sigmoid} function, so that the activation is scaled into $[0,1]$ and acts on the input layer by filter-wise multiplication. Different from \cite{hu2017} which sets this module at each residual layer, we only add it to transition layers in order to adjust the filters into the next block.

\noindent\textbf{Bottleneck and compression.} Bottleneck is an effective way to decrease the number of parameters and provide further potential to enlarge model capacity. It is conjectured \cite{zagoruyko2016wide} that bottleneck architecture is suitable for deeper networks and large dataset like ImageNet, and recent studies have embraced bottleneck for a better performance \cite{he2016deep,huang2016densely,Wang_2017_CVPR,chen2017dual}. So we introduce bottleneck to our large models. The $3\times3$ convolution kernels in each block are replaced by $1\times1$,
and produce a middle layer, after which, a $3\times3$ convolution layer follows to produce the top layer. The middle layer and top layer contain the same number of feature maps. Compression is another tool adopted in \cite{huang2016densely} to make the model more compact. Instead of compressing the number of filters in transition layers as they do, we only compress the features that are accessed to the loss function, \emph{i.e.} the Stage-II concatenated with its input layer. The models with compression have an extra convolutional layer with $1\times1$ kernel size before global pooling. It generates half the number of filters to enhance model compactness and keep the dimensionality of the final feature in a proper range.


\subsection{Implementation}

In our experiments, we test our models on benchmark datasets without the aforementioned extra techniques to show the effectiveness of CliqueNet, and further improve the state-of-the-art performance with them. There are two structure parameters, the sum of layers in all blocks, \emph{T}, and the number of filters per layer, \emph{k}. For our models without bottleneck, convolution layers in each block are with $3\times3$ kernel size and padded by one pixel to keep the feature maps in the same size. Blocks are linked by transition layers, where a convolution layer with $1\times1$ kernel size is followed by $2\times2$ average pooling. All convolutions are performed in a unit composed of three consecutive operations: batch normalization\cite{ioffe2015batch}, Relu, and the convolution. Stage-II feature with its input layer from all blocks are concatenated after global pooling, and end with a fully-connected layer with softmax.

\begin{table}[ht]
\begin{center}
\begin{tabular}{c|c|c|c|c}  
\hline
\small Layer & \small S0 & \small S1 & \small S2 & \small S3\\
\hline
\small Convolution & \multicolumn{4}{c}{\multirow{2}*{\small conv ($7\times7$), 64, stride 2}}\\
\small ($112\times112$) ~ \\
\hline
\small Pooling & \multicolumn{4}{c}{\multirow{2}*{\small max pool ($3\times3$), stride 2}}\\
\small ($56\times56$) ~ \\
\hline
\small Block 1 & \multirow{2}*{\small $36\times5$} & \multirow{2}*{\small $36\times5$} & \multirow{2}*{\small $36\times5$} & \multirow{2}*{\small $40\times6$}\\
\small ($56\times56$) & ~ & ~ & ~ & ~\\
\hline
\multicolumn{5}{c}{\small Transition: conv ($1\times1$), avg pool ($2\times2$)}\\
\hline
\small Block 2 & \multirow{2}*{\small $64\times6$} & \multirow{2}*{\small $80\times6$} & \multirow{2}*{\small $80\times5$} & \multirow{2}*{\small $80\times6$}\\
\small ($28\times28$) & ~ & ~ & ~ & ~\\
\hline
\multicolumn{5}{c}{\small Transition: conv ($1\times1$), avg pool ($2\times2$)}\\
\hline
\small Block 3 & \multirow{2}*{\small $100\times6$} & \multirow{2}*{\small $120\times6$} & \multirow{2}*{\small $150\times6$} & \multirow{2}*{\small $160\times6$}\\
\small ($14\times14$) & ~ & ~ & ~ & ~\\
\hline
\multicolumn{5}{c}{\small Transition: conv ($1\times1$), avg pool ($2\times2$)}\\
\hline
\small Block 4 & \multirow{2}*{\small $80\times6$} & \multirow{2}*{\small $100\times6$} & \multirow{2}*{\small $120\times6$} & \multirow{2}*{\small $160\times6$}\\
\small ($7\times7$) & ~ & ~ & ~ & ~\\
\hline
\end{tabular}
\end{center}
\caption{Structures on ImageNet. The first number in each block is the number of filters per layer, and the second denotes the number of layers in this block.}
\label{imagenet-structure}
\end{table}

For experiments on CIFAR and SVHN, there are three blocks in total, in which the feature map sizes are $32\times32$, $16\times16$, and $8\times8$, respectively. Before entering the first block, the input images pass through a $3\times3$ convolution with output channels set to be 64 as the input layer ($X_0$) of the first block. As for ImageNet, we use four blocks with bottleneck and compression, and compare our results with and without attentional transition. The initial transition has $7\times7$ convolution with stride 2 and $3\times3$ max pooling with stride 2 on the $224\times224$ input images. Our four network structures on ImageNet are shown in Table~\ref{imagenet-structure}.

\begin{table*}[!ht]
\begin{center}
\begin{tabular}{l|c c c|c|c|c|c|c}  
\hline
Model & A & B & C & FLOPs & Params & CIFAR-10 & CIFAR-100 & SVHN\\
\hline
Recurrent CNN \cite{liang2015recurrent} & - & - & - & - & 1.86M & 8.69 & 31.75 & 1.80\\
Stochastic Depth ResNet \cite{huang2016deep} & - & - & - & - & 1.7M & 11.66 & 37.8 & 1.75\\
dasNet \cite{stollenga2014deep} & - & - & - & - & - & 9.22 & 33.78 & - \\
FractalNet \cite{larsson2016fractalnet} & - & - & - & - & 38.6M & 7.33 & 28.2 & 1.87\\
\hline
DenseNet ($k=12, T=36$) \cite{huang2016densely} & - & - & - & 0.53G & 1.0M & 7.00 & 27.55 & 1.79\\
DenseNet ($k=12, T=96$) \cite{huang2016densely} & - & - & - & 3.54G & 7.0M & 5.77 & 23.79 & 1.67\\
DenseNet ($k=24, T=96$) \cite{huang2016densely} & - & - & - & 13.78G & 27.2M & 5.83 & 23.42 & 1.59\\
\hline
CliqueNet ($k=36, T=12$) & - & - & - & 0.91G & 0.94M & 5.93 & 27.32 & 1.77\\
CliqueNet ($k=64, T=15$) & - & - & - & 4.21G & 4.49M & 5.12 & 23.98 & 1.62\\
CliqueNet ($k=80, T=15$) & - & - & - & 6.45G & 6.94M & \textbf{5.10} & \textbf{23.32} & \textbf{1.56}\\
CliqueNet ($k=80, T=18$) & - & - & - & 9.45G & 10.14M & \textbf{{\color{red} 5.06}} & \textbf{{\color{red} 23.14}} & \textbf{\color{red} 1.51}\\
\hline\hline
DenseNet ($k=12, T=96$) \cite{huang2016densely} & - & \checkmark & \checkmark & 0.58G & 0.8M & 5.92 & 24.15 & 1.76\\
DenseNet ($k=24, T=246$) \cite{huang2016densely} & - & \checkmark & \checkmark & 10.84G & 15.3M & 5.19 & \textbf{{\color{red} 19.64}} & 1.74\\
\hline
CliqueNet ($k=36, T=12$) & \checkmark & - & - & 0.91G & 0.98M & 5.8 & 26.41 & -\\
CliqueNet ($k=36, T=12$) & - & - & \checkmark & 0.98G & 1.04M & 5.69 & 26.45 & -\\
CliqueNet ($k=36, T=12$) & \checkmark & - & \checkmark & 0.98G & 1.08M & 5.61 & 25.55 & 1.69\\
CliqueNet ($k=80, T=15$) & \checkmark & - & \checkmark & 6.88G & 8M & \textbf{5.17} & 22.78 & \textbf{{\color{red} 1.53}}\\
CliqueNet ($k=150, T=30$) & \checkmark & \checkmark & \checkmark & 8.49G & 10.02M & \textbf{{\color{red} 5.06}} & \textbf{21.83} & \textbf{1.64}\\
\hline
\end{tabular}
\end{center}
\caption{Error rates (\%) on CIFAR-10, CIFAR-100, and SVHN \textbf{without} any data augmentation. In CliqueNets and DenseNets, $k$ is the number of filters per layer, and $T$ is the total number of layers in three blocks. ``A, B, C'' represents attentional transition, bottleneck and compression, respectively. The FLOPs of DenseNets are calculated by ourselves.}
\label{cifar}
\end{table*}

\section{Experiments}

We evaluate the CliqueNet on benchmark classification datasets, including CIFAR-10, CIFAR-100, SVHN and ImageNet, and compare our results with the state of the arts.

\subsection{Datasets and Training Details}

\noindent\textbf{CIFAR.} The CIFAR-10 and CIFAR-100 datasets \cite{krizhevsky2009learning} are both $32\times32$ colored images. CIFAR-10 dataset consists of 60,000 images in 10 classes, with 6,000 images in each class. There are 50,000 images for training and 10,000 images for testing. CIFAR-100 dataset is similar to CIFAR-10 but has 100 classes, each of which contains 600 images. For data normalization, we preprocess the dataset  by subtracting the mean and dividing by the standard deviation.

\noindent\textbf{SVHN.} The Street View House Number (SVHN) \cite{netzer2011reading} dataset contains $32\times32$ colored images of house numbers cropped from Google Street View. There are 73,257 images in the training set, 26,032 in the testing set and 531,131 digits for additional training. Following the common practice \cite{zagoruyko2016wide,huang2016deep,larsson2016fractalnet,huang2016densely}, we use all training samples without augmentation and divide images by 255 for normalization. We report the lowest error rate on the testing set.

\noindent\textbf{ImageNet.} We also conduct experiments on ILSVRC 2012 dataset\cite{deng2009imagenet}, which contains 1.2 million training images, 50,000 validation images, and 100,000 test images with 1,000 classes. Following \cite{he2016deep,huang2016densely}, we adopt the standard data augmentation for the training sets. A $224\times224$ crop is randomly sampled from the images or its horizontal flip. The images are normalized into $[0,1]$ using mean values and standard deviations. We report the single-crop error rate on the validation set.



\noindent\textbf{Training Details.} For fair comparison, we do not take much hyper-parameter tuning, and most of our training strategies are following \cite{he2016deep,huang2016densely}. We train our models using stochastic gradient descent (SGD) with 0.9 Nesterov momentum and $10^{-4}$ weight decay. The parameters are initialized according to \cite{he2015delving} and the weights of fully connected layer are using Xavier initialization \cite{glorot2010understanding}. For CIFAR and SVHN, we train for 300 epochs and 40 epochs, respectively, with batchsize of 64. The learning rate is set to be 0.1 initially and is divided by 10 at 50\% and 75\% of the training procedure. Compared with ImageNet, the experiments on CIFAR and SVHN are not resorting to any data augmentation, and we add a dropout layer \cite{srivastava2014dropout} with drop out rate 0.2 after each convolution layer following\cite{huang2016densely}. For ImageNet, we train our models for 100 epochs and drop the learning rate by 0.1 at epoch 30, 60, and 90. Because we have only server with 4 GPUs and are constrained by GPU memory, the batchsize is 160 for our models on ImageNet, instead of 256 as most studies did.




\begin{table}[ht]
\begin{center}
\begin{tabular}{l l c c}  
\hline
Model & Params & top-1 & top-5 \\
\hline
ResNet-18 \cite{he2016deep} & 11.7M & 30.43 & 10.76\\
CliqueNet-S0$^*$ & 5.7M &  27.52 &  8.98\\
\hline
ResNet-34 \cite{he2016deep} & 21.8M & 26.73 &  8.74 \\
CliqueNet-S1$^*$ & 7.96M & 26.21 & 8.3 \\
CliqueNet-S2$^*$ & 10M & 25.85 & 8.02 \\
\hline
DenseNet-121 \cite{huang2016densely} & 7.98M & 25.02 & 7.71 \\
CliqueNet-S2 & 11M & 24.82 & 7.51 \\
CliqueNet-S3$^*$ & 13.17M & 24.98 & 7.48\\
\hline
ResNet-50 \cite{he2016deep} & 25.6M & 24.01 & 7.02 \\
CliqueNet-S3 & 14.38M & 24.01 & 7.15 \\
\hline
\end{tabular}
\end{center}
\caption{Single crop error rates (\%) on ImageNet. The $*$ indicates the models without attentional transition.}
\label{imagenet}
\end{table}

\subsection{Results on CIFAR and SVHN}

Our experimental results on CIFAR and SVHN are shown in Table~\ref{cifar}. The first part in the table includes some methods before DenseNets and some other studies that also incorporate feedback connections or attention mechanism. The second and third parts compare the CliqueNets with DenseNets when they both have no extra technique. The last two parts show the situation with extra techniques. The best result and the second best result are marked by red bold and bold, respectively.

\noindent\textbf{Without extra techniques.} The first three parts show that, when extra techniques are not considered, CliqueNets outperform most previous methods on CIFAR-10, CIFAR-100, and SVHN with significantly fewer parameters. Because the layers in CliqueNet can be re-updated but contribute features in each cycle, the depth of CliqueNet is much shallower than other models. For our smallest model CliqueNet (36-12), (representing $k=36$, and $T=12$), each block contains 4 layers. It has the same number of filters, 144, in each block as DenseNet (12-36), but reduce the error rate from 7\% to 5.93\% on CIFAR-10 with slightly fewer parameters than its counterpart DenseNet (12-36). Although the ResNet with stochastic depth \cite{huang2016deep} achieved a slightly better performance with 1.7M parameters on SVHN than CliqueNet (36-12), our model drops the error rate on CIFAR-10 and CIFAR-100 by a large margin. As the model capacity goes larger, we find that the performance of CliqieNets is getting better without overfitting. As for our model CliqueNet (80-15), it has already achieved the state of the art on three datasets, and even outperforms the DenseNets that use extra techniques on CIFAR-10 and SVHN. It has only 6.94M parameters, which are a quarter of DenseNet (24-96) with 27.2M parameters, and a half of DenseNet (24-246) using bottleneck and compression with 15.3M parameters.

\noindent\textbf{With extra techniques.} The CliqueNets realize spatial attention mechanism due to its recurrent feedback propagation. When armed with channel-wise attention, they achieve an improved performance. This is demonstrated by the CliqueNet (36-12) with attentional transition. It has a better result on CIFAR-10 and CIFAR-100 with slightly more parameters. The compression has the same effect by making the model more compact. It is shown that the attentional transition is compatible with compression. The CliqueNet (36-12) with both attentional transition and compression leads to a better result than its original version and its original version with only attentional transition or compression. Compared with its counterpart DenseNet (12-36), it drops an error rate of 1.39\% on CIFAR-10, 2\% on CIFAR-100, and 0.1\% on SVHN, with just 0.08M more parameters. The CliqueNet (80-15) with attentional transition and compression also has an improvement than its original version, and increases the state of the art of SVHN to 1.53\% with 8M parameters, while the previously best result 1.59\% on SVHN performed by DenseNet (24-96) has three times more parameters. The bottleneck architecture is effective to save parameters, and our largest model CliqueNet (150-15) with bottleneck further improves the performance on CIFAR-10 and CIFAR-100, but increases parameter and computation cost moderately.

\begin{figure}[!t]
\begin{center}
   \includegraphics[width=1\linewidth]{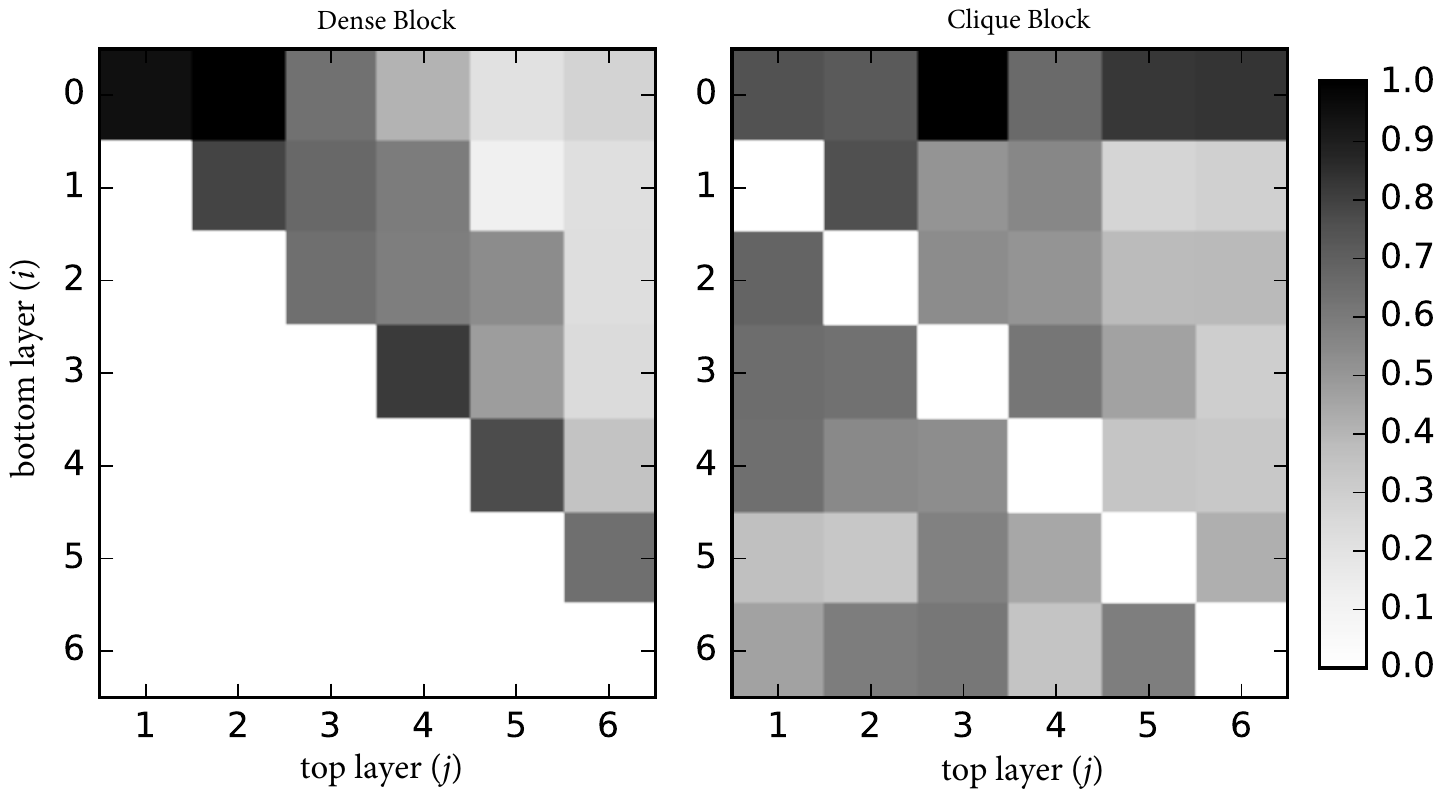}
\end{center}
   \caption{Visualization of the weights in the first block in pre-trained DenseNet (left) and CliqueNet (right) by calculating the average absolute value of $W_{ij}$. Node 0 denotes the input layer of this block.}
\label{weight-heatmap}
\end{figure}

\subsection{Results on ImageNet}

Because we have limited computational resource and can only spread a batch among 4 GPUs, we use a batchsize of 160 on ImageNet, instead of 256 in most studies. Although a smaller batchsize would impair the performance training for the same epochs, the CliqueNets achieve a comparable result on ImageNet with ResNets or DenseNets; see Table~\ref{imagenet}. This indicates that our proposed models can also be applied on large datasets.

The CliqueNet-S0$^*$ and CliqueNet-S1$^*$ outperform the ResNet-18 and ResNet-34 with only a half of their parameters. Larger models also achieve on par with the state of the art performed by ResNets and DenseNets. When the attentional transition is considered, the CliqueNet contains both spatial attention and channel-wise attention, and has a better performance accordingly. The CliqueNet-S2 and CliqueNet-S3 both reduce about 1\% top-1 error rate compared with their original versions, CliqueNet-S2$^*$ and CliqueNet-S3$^*$ that do not have attentional transition.

\subsection{Further Discussion}

In order to better analyze the recurrent feedback mechanism and the multi-scale feature strategy in CliqueNet, we visualize feature maps and parameters based on pre-trained models and provide a further understanding.

\noindent\textbf{Parameter efficiency.} Despite the fact that CliqueNet has bipartite connections between any two layers in the same block, which would bring more parameters in the block, we find that the CliqueNet achieves the state of the art on CIFAR and SVHN dataset with considerably fewer parameters than DenseNets. On ImageNet, the CliqueNet using a smaller batchsize also has parameter efficiency compared with ResNets. This is mainly due to the multi-scale feature strategy that only transits the Stage-II feature into the next block, instead of having feature maps stacked towards deeper layers, which may cause progressive increment of parameters. In Figure~\ref{weight-heatmap}, we visualize the weights among layers within a block of pre-trained CliqueNet and DenseNet. The color pixel of Clique Block covers the whole heat map because of our feedback connections. It is noted that the heat dots in a Dense Block are concentrated along the diagonal. A similar result is also reported in \cite{huang2016densely}. The observation reveals that only neighboring layers have strong dependency in DenseNet, while its forward stacking pattern is actually parameter-demanding. This helps to explain the parameter and flop efficiency in CliqueNet where information flow is distributed more evenly in each block.

\begin{figure}[!ht]
\begin{center}
   \includegraphics[width=1\linewidth]{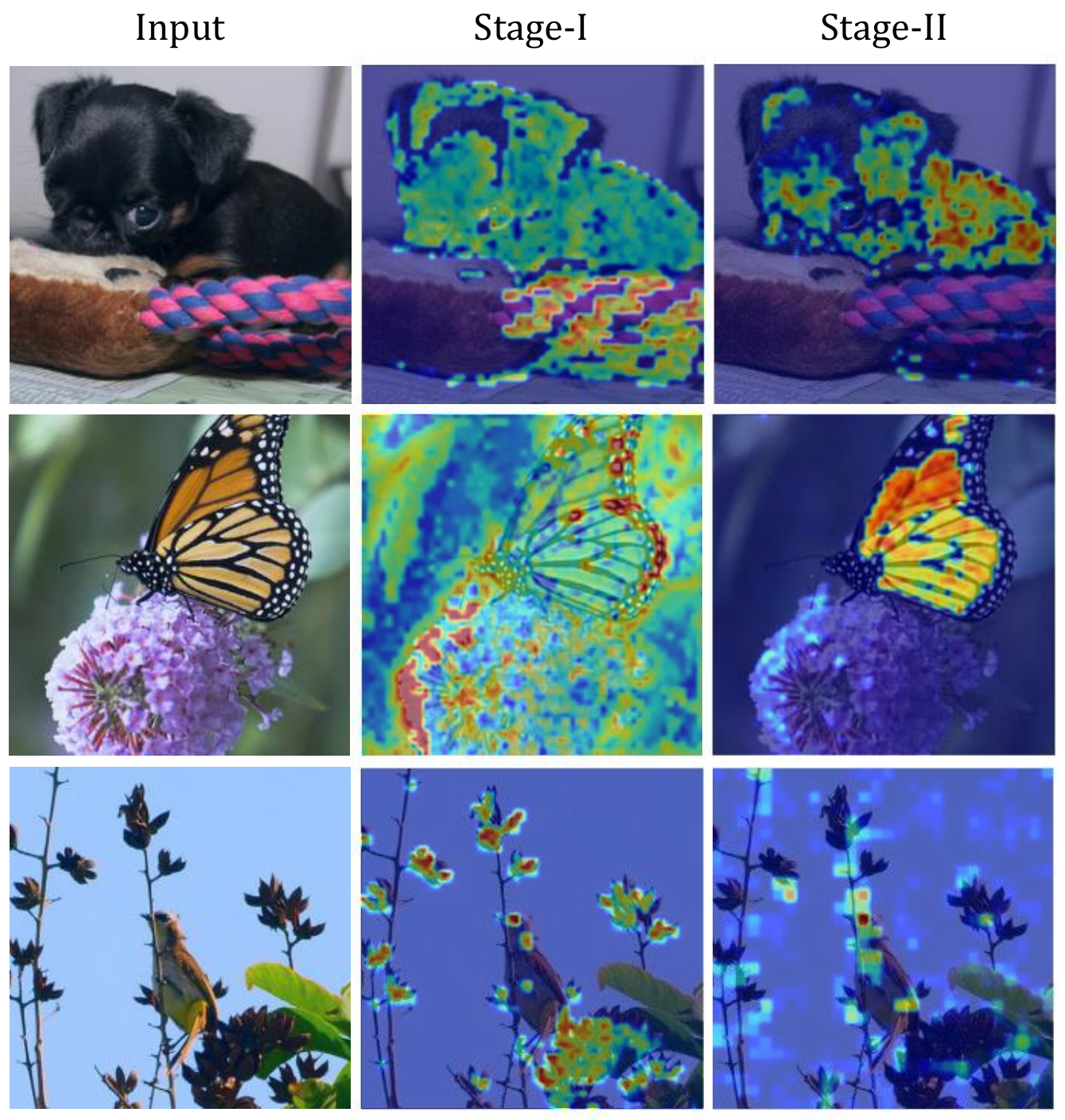}
\end{center}
   \caption{Feature maps of Stage-I and Stage-II with the highest average activation in a pre-trained model. The activations of background or surrounding objects are repressed in Stage-II.}
\label{data-heatmap}
\end{figure}

\noindent\textbf{Feature refinement.} In CliqueNet, the layers are updated alternately so that they are supervised by each other. Moreover, in the second stage, feature maps always receive a higher-level information from the filters that are updated more lately. This spatial attention mechanism makes layers refined repeatedly, and is able to repress the noises or background of images and focus more activations on the region that characterize the target object. In order to test the effects, we visualize the feature maps following the methods in \cite{zeiler2014visualizing}. As shown in Figure~\ref{data-heatmap}, we choose three input images with complex background from ImageNet validation set, and visualize their feature maps with the highest average activation magnitude in the Stage-I and Stage-II, respectively. It is observed that, compared with the Stage-I, the feature maps in Stage-II diminish the activations of surrounding objects and focus more attention on the target region. This is in line with the conclusion in Table~\ref{allnet-L} that the Stage-II feature is more discriminative and leads to a better performance.

\section{Conclusion}

In this study, we introduce a new convolutional neural network architecture where the layers in a block are constructed as a clique and are updated alternately in a loop manner. Any layer is both the input and output of another one in the same block so that the information flow is maximized. The parameters are circulated in the course of propagation and are able to produce multiple stage features. We analyze the feature in different stages and observe that the introduce of the Stage-II feature helps to suppress noises and leads to a better performance. The multi-scale feature strategy effectively circumvents the progressive increment of parameters. Experiments show that our proposed architectures are able to achieve the state of the arts with fewer parameters, especially on CIFAR and SVHN without resorting to data augmentation.

Different from prior networks, the CliqueNet utilizes a fixed number of parameters to attain a deeper representation space and incorporates the recurrent feedback to achieve attention mechanism. This topology provides the potential of developing models for other computer vision tasks in future work, such as semantic segmentation, salient object detection, image captioning, \emph{etc.}

\section*{Acknowledgements}

Zhouchen Lin was supported by National Basic Research Program of China (973 Program) (grant no. 2015CB352502), National Natural Science Foundation (NSF) of China (grant nos. 61625301 and 61731018), Qualcomm, and Microsoft Research Asia.

{\small
\bibliographystyle{ieee}
\bibliography{egbib}
}

\end{document}